# CIEGAD: Cluster-Conditioned Interpolative and Extrapolative Framework for Geometry-Aware and Domain-Aligned Data Augmentation

Keito Inoshita[1], Xiaokang Zhou[1], Akira Kawai[2], Katsutoshi Yada[1]
[1]Kansai University, Japan
[2]Shiga University, Japan
Email: inosita.28652@gmail.com

*Abstract*—In practical deep learning deployment, the scarcity of data and the imbalance of label distributions often lead to semantically uncovered regions within the real-world data distribution, hindering model training and causing misclassification near class boundaries as well as unstable behaviors in peripheral areas. Although recent large language models (LLMs) show promise for data augmentation, an integrated framework that simultaneously achieves directional control of generation, domain alignment, and quality control has not yet been fully established. To address these challenges, we propose a Cluster-conditioned Interpolative and Extrapolative framework for Geometry-Aware and Domain-aligned data augmentation (CIEGAD), which systematically complements both in-distribution and out-of-distribution semantically uncovered regions. CIEGAD constructs domain profiles through cluster conditioning, allocates generation with a hierarchical frequency–geometric allocation integrating class frequency and geometric indicators, and finely controls generation directions via the coexistence of interpolative and extrapolative synthesis. It further performs quality control through geometry-constrained filtering combined with an LLM-as-a-Judge mechanism. Experiments on multiple classification tasks demonstrate that CIEGAD effectively extends the periphery of real-world data distributions while maintaining high alignment between generated and real-world data as well as semantic diversity. In particular, for long-tailed and multi-class classification tasks, CIEGAD consistently improves F1 and recall, validating the triple harmony of distributional consistency, diversity, and quality. These results indicate that CIEGAD serves as a practically oriented data augmentation framework that complements underrepresented regions while preserving alignment with real-world data.

## I. INTRODUCTION

THE widespread adoption of deep learning has significantly advanced both the performance and real-world applicability of various intelligent systems, including emotion recognition, anomaly detection, recommendation, and dialogue support. The high accuracy and robustness of these models largely rely on the availability of sufficient amounts of carefully curated training data, which enables the empirical learning of domain-specific distributional characteristics such as lexical complexity, stylistic variations, topic transitions, and ambiguity near class boundaries [1]. However, in practical and research settings, such well-prepared large-scale datasets are not always available. In many cases, data collection is costly, time-consuming, and constrained by legal or domain-specific limitations, resulting in limited data or class imbalance issues [2]. Under these conditions, the distributional contour becomes indistinct, and rare expressions or pragmatic variations in peripheral regions are often missing. Consequently, models trained on existing data are forced to perform extrapolative reasoning in unseen regions, showing good performance around central regions but frequently misclassifying samples near boundaries or in outlying areas. Addressing this issue requires not only enhancing model capacity but also adopting data augmentation approaches that compensate for missing distributional knowledge [3].

In this context, large language models (LLMs) are highly promising because they can generate coherent and diverse text conditioned on a given context and constraints, leveraging linguistic knowledge acquired from large-scale corpora. Beyond simple paraphrasing, LLMs are capable of producing continuous variants of domain-specific expressions, rhetorical structures, and degrees of implication, while subtly altering aspects such as speaker, situation, or topical focus. This makes them well-suited for data augmentation tasks [4]. By harnessing the generative capabilities of LLMs, it becomes possible to augment locally sparse regions in the data space affected by scarcity and imbalance, enabling strategic expansion of target domain distributions while maintaining structural consistency, rather than merely increasing data volume. Such augmentation can substantially enhance the accuracy and robustness of downstream tasks.

Nevertheless, existing LLM-based augmentation methods tend to emphasize the diversity of generated text through persona assignment or prompt engineering, while providing limited control over the direction of augmentation. As a result, generated samples often drift toward general linguistic phenomena, deviating from the target domain [5]. Conversely, approaches that rely heavily on real-world data maintain strong domain alignment but often fail to sufficiently cover semantically uncovered regions, areas within or around the data distribution that are underrepresented in meaning, leading to insufficient extrapolation [6]. In particular, there is a lack of mechanisms that ensure extrapolative augmentation for both in-distribution and out-of-distribution robustness while preserving domain alignment. Moreover, the aspect of quality control in generated data has been largely overlooked. Without adequate quality control, noise in the form of redundancy, contradiction, or unnatural implication can be introduced, resulting in long-term degradation of downstream task performance and the accumulation of synthetic data debt [7]. When performing extrapolative augmentation, poor-quality samples pose further risks, as unnatural language can propagate beyond the original domain. Therefore, it is essential to establish an integrated pipeline that maintains domain alignment, systematically augments semantically uncovered regions both inside and outside the domain distribution, and enforces strict quality control at the acceptance stage.

To address these challenges, this study proposes a Cluster-conditioned Interpolative and Extrapolative framework for Geometry-Aware and Domain-aligned data augmentation (CIEGAD). CIEGAD systematically complements semantically uncovered regions while maintaining alignment with real-world data distributions through both interpolative and extrapolative data generation. Specifically, the framework dynamically determines the number of clusters for each class based on the number of samples and constructs a domain profile for each cluster composed of core and periphery examples. This enables conditional generation that captures local domain-specific features that cannot be represented by simple class conditioning. During generation allocation, CIEGAD employs a Hierarchical Frequency–Geometric Allocation (HFGA) that operates across both the class and cluster levels. At the class level, inverse-frequency weighting is applied to compensate for long-tailed classes, while at the cluster level, generation proportions are further refined by integrating inverse cluster frequency, inter-cluster geometric distance, and local density. This mechanism allows the framework to strategically expand small, peripheral, and sparse regions within the data distribution. During data synthesis, CIEGAD combines interpolation and extrapolation based on the embedding distribution of each cluster, where interpolation smoothly complements semantically uncovered areas within the cluster, and extrapolation controllably extends the peripheral boundary. For quality control, geometry-based and similarity-based filters automatically eliminate redundant or inconsistent samples, while an LLM-as-a-Judge module evaluates semantic attributes such as emotion consistency, style alignment, and diversity, accepting only samples that meet a defined threshold. By iteratively executing these processes, CIEGAD expands semantically uncovered regions in a planned and controlled manner while preventing distributional deviation.

The main contributions of this paper are summarized as follows:

i) An integrated data augmentation framework, CIEGAD, is developed to unify the processes of data characterization, data generation, and quality control, systematically complementing semantically uncovered regions within and beyond real-world data distributions while maintaining domain alignment.

ii) A cluster-profile-driven generation strategy is designed based on a HFGA, enabling domain-aligned synthesis that adaptively reinforces long-tailed classes and locally structured regions.

iii) A joint mechanism of generation control and quality control is realized by integrating interpolative and extrapolative data generation with geometric filtering and LLM-as-a-Judge evaluation, achieving high-quality semantic diversity while preserving distributional structure.

The rest of this paper is organized as follows. Section II reviews related work and positions our study within existing research. Section III presents the detailed framework of CIEGAD. Section IV describes the experimental design. Section V reports experimental results and provides analysis. Section VI discusses the key findings and limitations. Finally, Section VII concludes the paper.

## II. Related Work

### A. Data Augmentation for Model Performance Improvement

Text data augmentation has long been studied as an essential technique to improve model generalization and facilitate learning under low-resource conditions. Early studies mainly relied on simple rule-based approaches, such as word insertion, substitution, and deletion. Wei and Zou [8] proposed the representative method easy data augmentation (EDA), which achieved performance gains through surface-level textual transformations. However, such rule-based methods often introduce semantic distortion and noise, motivating the need for more precise and controllable augmentation approaches.

Recently, advanced generative techniques based on LLMs have rapidly gained popularity. Zhao et al. [9] compared two approaches, rewriting existing sentences and generating entirely new ones using ChatGPT, and found that novel generation tends to yield better classification performance, although its effectiveness strongly depends on prompt design and generation quantity. Becker et al. [10] focused on the reliability of assigning cs to LLM-generated data and applied conditional label smoothing in clinical text classification, which effectively reduced false negatives caused by label noise. Yang et al. [11] proposed Mini-DA, a selective augmentation approach that targets only difficult samples, thereby achieving efficient improvement in classification accuracy.

Several foundational frameworks have also contributed to this direction. Balkus and Yan [12] verified GPT-3–based augmentation methods, while Dai et al. [13] proposed AugGPT, which employs ChatGPT to generate diverse yet faithful text samples. AugGPT emphasized the importance of balancing fidelity and diversity, as managing the trade-off between quality and variety is crucial for augmentation effectiveness. Moreover, Chai et al. [14] conducted a comprehensive survey of LLM-based data augmentation and pointed out that most existing methods fail to consider the directionality of augmentation or the alignment between generated and real-world data distributions.

Research has also addressed class imbalance problems. Zhang et al. [15] introduced a SMOTE-inspired [16] approach in which representative samples are interpolated through LLM-based synthesis to balance long-tailed distributions, significantly improving F1 and macro-accuracy scores. Collectively, these studies have explored various perspectives of LLM-based data augmentation, including promoting diversity, ensuring label reliability, and handling rare classes. However, existing methods still lack structural understanding of overall data distributions and fail to geometrically complement semantically uncovered regions, leading to insufficient control over the direction of augmentation. To address this gap, the proposed CIEGAD introduces interpolative and extrapolative data generation guided by cluster structures, achieving unified control of both distributional alignment and semantic diversity.

### B. Domain Alignment for Consistent Data Generation

In text generation models, maintaining domain-specific characteristics consistent with the target task or application context is crucial. While the generality and flexibility of LLMs are valuable, it is equally important to guide and constrain their outputs to reflect domain-specific knowledge, vocabulary, and stylistic traits. Wang et al. [17] designed structured prompts that explicitly define roles, purposes, and tasks to maintain alignment with clinical guidelines, achieving domain-consistent outputs. Similarly, Liu et al. [18] improved generation quality in materials science by presenting structured information, such as material names, properties, and applications, as text prompts for LLMs. Feng et al. [19] enhanced arrhythmia diagnosis from electronic health records by designing explanation-based prompts that encouraged structured, interpretable outputs, substantially improving diagnostic accuracy. These studies indicate that providing detailed domain-defining information helps ensure alignment and control in generated outputs.

Frameworks that systematically support prompt engineering have also emerged. Morales et al. [20] proposed a domain-specific language–based framework that enables prompt structuring, reuse, and validation, establishing a foundation for domain-aligned generation. Moreover, prompt structure itself can be learned as part of the model parameters. Zhou et al. [21] introduced CoOp, a framework that incorporates learnable context vectors into vision–language models, significantly enhancing domain adaptation performance. Similarly, Wei et al. [22] proposed a prompt-learning approach for image captioning that jointly optimizes attribute consistency and semantic consistency, improving generalization to unseen domains.

These approaches share a lightweight and flexible design that maintains domain alignment without modifying the underlying LLM. However, they often struggle to capture fine-grained variations within the same domain, such as local clusters or stylistic subdomains. To address this limitation, CIEGAD automatically extracts clusters for each label and constructs domain profiles that condition the generation process, thereby preserving domain alignment at a local level.

### C. Diversity Enhancement for Complementary Data Generation

Diversity is a fundamental factor in text data augmentation. Diverse data not only prevent model overfitting and improve generalization but also expand coverage of long-tailed classes and localized clusters. However, generated outputs often gravitate toward dominant distributions or template-like structures, producing redundant results that lack meaningful variation. Consequently, recent studies have increasingly explored prompt design and output control mechanisms to encourage diversity in LLM-based generation.

From a prompt-engineering perspective, in-context learning has become a promising approach to promote diversity. Ye et al. [23] proposed CEIL, which selects in-context examples by balancing relevance and diversity using determinantal point processes (DPP), achieving more stable and generalized results. Yang et al. [24] extended this idea with a two-stage DPP approach that removes redundancy between examples while preserving diversity and quality. Kapuriya et al. [25] addressed the limitations of similarity-based selection by employing maximum marginal relevance to diversify topic coverage.

From the perspective of cognitive or cultural diversity, Hayati et al. [26] introduced a staged prompting strategy inspired by human opinion-formation processes, encouraging varied perspectives in generated text. Similarly, Wang et al. [27] proposed multilingual prompts to activate multicultural knowledge within LLMs, achieving lexical and conceptual diversity. Lau et al. [28] developed Dipper, which aggregates outputs from multiple reasoning-style prompts to exploit ensemble-like inference diversity within a single model. These approaches are attractive for practical use since they enhance diversity without additional model training.

On the other hand, they typically lack mechanisms to control the direction of diversity, meaning the expansion occurs without geometric or semantic constraints. Reinforcement learning–based approaches [29] have been proposed to directly control generation diversity, providing finer control but requiring custom reward design and additional model training, which increases deployment cost and complexity. Considering these limitations, practical data augmentation requires a framework that can balance diversity and quality without retraining. The proposed CIEGAD achieves this by geometrically defining generation directions and integrating quality control, providing a lightweight yet effective augmentation paradigm suitable for real-world applications.

## III. CIEGAD FOR DATA AUGMENTATION INTO SEMANTICALLY UNCOVERED REGIONS

### A. Framework Overview

Real-world data often exhibit biased lexical and stylistic distributions, and performance tends to degrade sharply in semantically uncovered regions, especially for tasks involving emotion or intent understanding. Conventional data augmentation methods typically address sparsely populated regions within the distribution through interpolation, thereby improving generalization to unseen contexts. However, their robustness to outliers and out-of-distribution phenomena, which frequently occur in real-world applications, remains limited. To achieve practically effective data augmentation, it is essential to expand data in a manner that is both diverse and aligned with the intrinsic structure of real-world data, while introducing extrapolative mechanisms to enhance robustness against out-of-distribution instances.

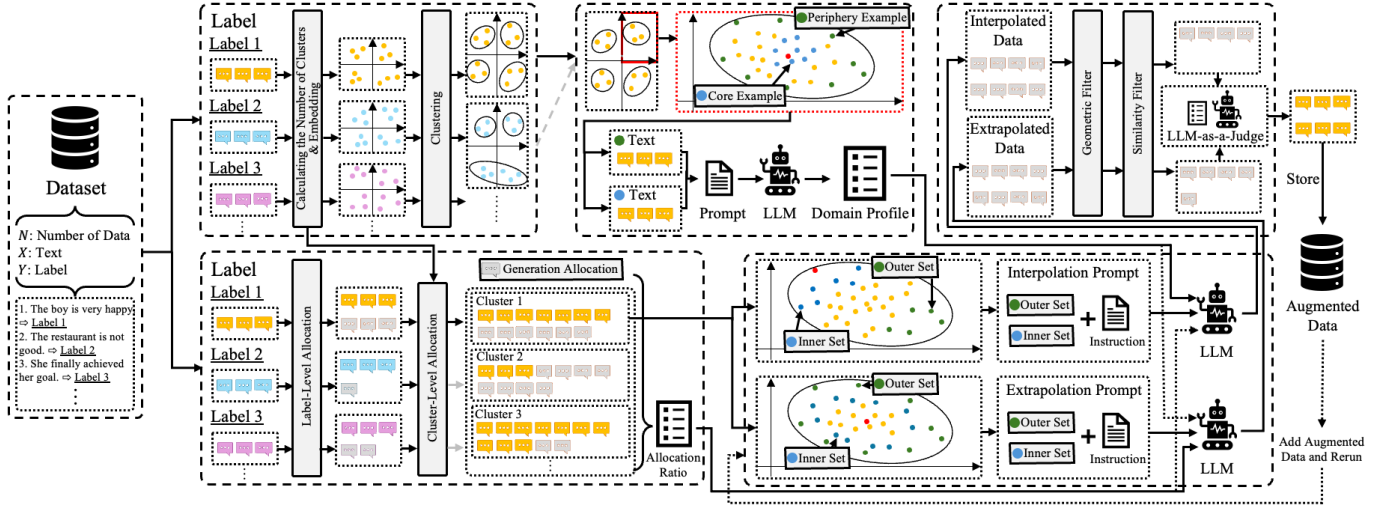

Fig. 1. Framework Architecture of CIEGAD.

To this end, this study proposes CIEGAD, a framework that systematically complements semantically uncovered regions both inside and outside the real-world data distribution through cluster-conditioned interpolative and extrapolative data generation. The overall architecture of CIEGAD is illustrated in Fig. 1. For each class-label $y \in \{1, 2, ..., L\}$, CIEGAD determines the number of clusters based on the number of samples belonging to that class. The real-world data are embedded to obtain representations $x_{y,i} \in R^d$, and the data for each class are dynamically partitioned into clusters using k-means clustering with the determined cluster number. Each cluster is regarded as a local domain that shares distinctive lexical, stylistic, topical, and affective mixtures within the same class-label. For each cluster, a domain profile is generated to describe its linguistic and contextual characteristics, serving as a conditioning basis for maintaining domain alignment during data generation. Additionally, a HFGA is applied to determine the number of augmented samples per cluster, based on the characteristics of each class and its constituent clusters. HFGA allows the total augmentation target $T$ to be appropriately distributed, prioritizing clusters with lower diversity or insufficient coverage.

Under these conditions, the data generation phase defines the inner and outer sets of each cluster in the embedding space. Interpolative generation fills semantically uncovered regions within the cluster, while extrapolative generation expands the outer boundaries to complement semantically uncovered regions beyond the existing data distribution. All generated samples are then filtered through geometric constraint filters and similarity filters, followed by an evaluation using LLM-as-a-Judge [30], which assesses linguistic naturalness. Only samples that pass all filters are accepted as augmented data for the respective clusters, after which the cluster datasets are updated. This process of inner–outer identification, generation, and evaluation is iteratively repeated until the augmentation target $T$ is achieved. Through this design, CIEGAD preserves the advantages of traditional data augmentation while improving robustness to outliers and out-of-distribution phenomena commonly observed in real-world applications.

*B. Domain Profile for Cluster Conditioning*

In data augmentation, it is critical not only to generate additional samples but also to ensure domain alignment between real-world and generated data. Understanding the detailed characteristics of the original data enables contextually appropriate generation within the real-world data distribution. Therefore, CIEGAD dynamically determines the number of clusters per class-label $y$ and constructs a domain profile for each cluster to capture fine-grained domain-specific properties. The processing flow is illustrated in Fig. 2. The resulting domain profiles serve as cluster-level conditioning information to maintain domain alignment during generation.

The dynamic determination of the number of clusters is essential to capture multimodal structures within each class-label and to obtain a detailed cluster-level understanding when constructing domain profiles. Specifically, the number of clusters $K_y$ for each class $y$ is determined based on the number of samples $N_y$ as follows:

$$K_y = \min\left(\left\lceil\sqrt{\frac{N_y}{\kappa}}\right\rceil + 2, high\right) \quad (1)$$

where the min function prevents excessively large or overly small cluster numbers. In this study, $high = 18$ is set, allowing a decomposition into at most 18 clusters and at least 3. A scaling coefficient $\kappa = 800$ is used, determined empirically to balance the representational granularity of topic and stylistic diversity within each class-label. Consequently, classes with fewer samples produce coarser clusters that capture representative styles and core discourse patterns, while larger classes are decomposed into finer clusters that distinguish subtle lexical variations, stylistic tendencies, and differences in emotional intensity. This adaptive mechanism prevents both over-segmentation and under-segmentation, yielding an optimal number of clusters per class.

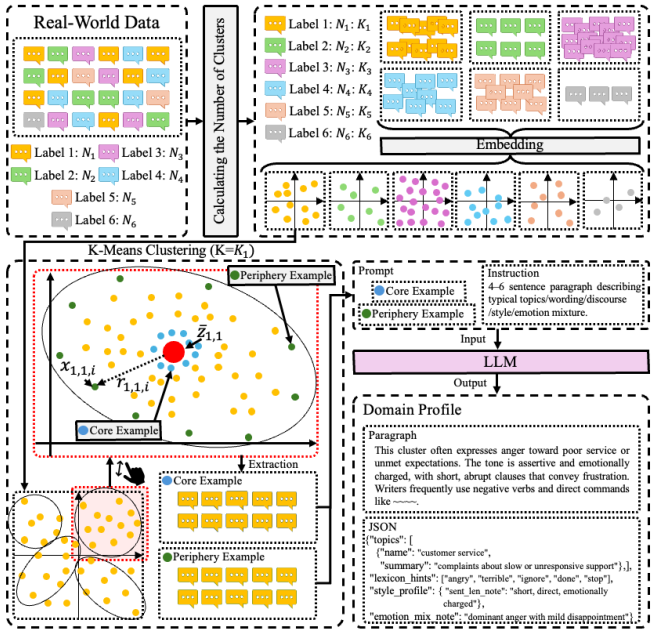

Fig. 2. Mechanism of Domain Profile Construction.

Next, for each cluster, a domain profile describing its detailed characteristics is constructed. Each dataset corresponding to class $y$ is first embedded using Sentence Transformer [31], resulting in $X_y$. k-means clustering is then performed using the previously determined number of clusters, and for each cluster $k \in \{1, 2, \ldots, K_y\}$, the corresponding data subset $X_{y,k} = \{x_{y,k,1}, x_{y,k,2}, \ldots, x_{y,k,i}\}$ is obtained. Based on the Euclidean distance $r_{y,k,i} = \|x_{y,k,i} - \bar{z}_{y,k}\|_2$ from the cluster centroid $\bar{z}_{y,k}$, the 10 samples closest to the centroid are selected as core examples, and the 10 farthest samples are selected as periphery examples. These examples define the cluster's center and periphery, providing a structured and summarized representation of the local domain.

To construct the domain profile, a prompt containing an instruction, core examples, and periphery examples is provided to the LLM, from which two layers of information are extracted:

- Cluster-level summary: extracts background knowledge of stylistic and rhetorical characteristics from the core and periphery examples, defining the range of augmentation grounded in cluster properties.
- Structured information such as topics and frequent expressions: extracts salient topics and frequently used expressions to provide explicit domain-level information directly usable by the LLM during generation.

The resulting domain profiles function as conditioning inputs to the LLM during data generation, preserving cluster-specific linguistic features and writing styles. This enables effective data augmentation in a manner consistent with the underlying domain structure.

*C. HFGA for Data Imbalance*

In data augmentation, naively generating samples does not address class-level imbalance inherent in real-world data. Moreover, multiple clusters often exist within each class, where inter-cluster imbalance further complicates diversity expansion. CIEGAD therefore adopts a HFGA that determines per-cluster generation counts by jointly considering the class level and the cluster level. Let the total number of original samples be $N$ and the augmentation ratio be $\rho$. The target number of accepted augmented samples is $T = \lfloor \rho N \rfloor$.

First, let $N_y$ be the number of samples in class $y$. To favor minority classes, we compute an inverse-frequency weight with a small $\varepsilon > 0$ to avoid division by zero, as follows:

$$w_y = \frac{1}{N_y + \varepsilon} \quad (2)$$

The class-level allocation $A_y$ is then obtained by normalizing these weights so that the total allocation equals $T$:

$$A_y = round\left(T \cdot \frac{w_y}{\sum_{y'} w_{y'}}\right) \quad (3)$$

Next, we distribute $A_y$ across clusters $k$ within class $y$. To do so, we introduce a cluster priority score $v_{y,k}$, which quantifies the priority of data generation for each cluster. This score is defined as a linear combination of three geometric and statistical factors, cluster size, inter-cluster separation, and intra-cluster sparsity, as follows:

$$v_{y,k} = \alpha \cdot \frac{1}{N_{y,k} + 10^{-6}} + \beta \cdot \min_{k' \neq k} \|\bar{z}_{y,k} - \bar{z}_{y,k'}\|_2$$
$$+ \gamma \cdot \mathbb{E}_{i \in X_{y,k}}\left[\sqrt{2(1 - \cos(x_{y,k,i}, \bar{z}_{y,k}))}\right] \quad (4)$$

where $N_{y,k}$ denotes the size of cluster $k$ within class $y$, and $\bar{z}_{y,k}$ represents the centroid of cluster $k$. The first term assigns larger weights to smaller clusters, emphasizing low-sample clusters. The second term represents the minimum Euclidean distance to other clusters, giving higher priority to geometrically isolated, peripheral clusters that are underrepresented in the embedding space. The third term approximates the local sparsity of each cluster by averaging the cosine-distance displacement of samples from the centroid, converted to Euclidean distance via $\sqrt{2(1 - \cos(\cdot))}$. As a result, clusters with lower internal density obtain higher sparsity scores.

In this study, we set $(\alpha, \beta, \gamma) = (0.5, 0.25, 0.25)$ to balance the three aspects, small-scale, peripheral, and sparse structures, providing an empirical equilibrium that primarily corrects class imbalance while moderately reflecting geometric diversity within the data distribution.

Finally, the normalized cluster-level allocation $A_{y,k}$ is calculated based on $v_{y,k}$ as follows:

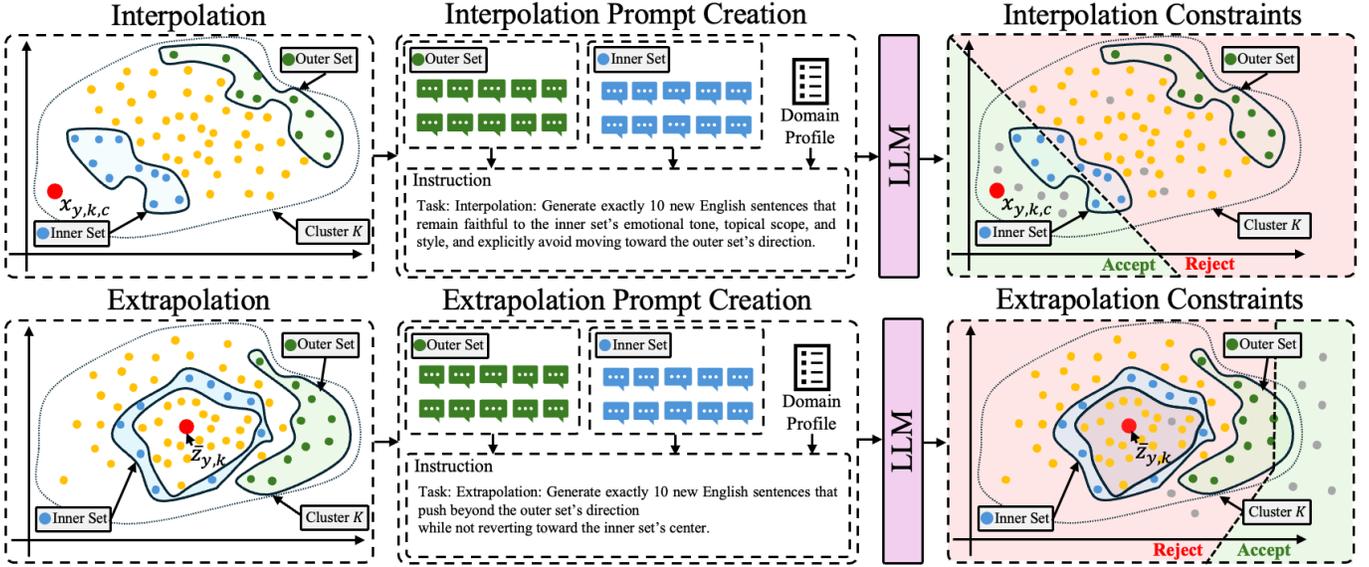

Fig. 3. Mechanism of Interpolative and Extrapolative Data Generation.

$$A_{y,k} = \text{round}\left(A_y \cdot \frac{v_{y,k}}{\sum_{k'} v_{y,k'}}\right) \quad (5)$$

Through this hierarchical allocation process, generation resources are concentrated on small, peripheral, and sparse clusters, effectively complementing gaps in the data distribution. This design enables CIEGAD to determine an appropriate generation allocation that considers both class- and cluster-level imbalance, leading to more balanced and representative data augmentation. The complete algorithmic pipeline is illustrated in Algorithm 1.

### D. Interpolative and Extrapolative Data Generation for Improving Robustness

To enable data augmentation that is truly effective in real-world applications, it is essential to expand data in a manner that is both diverse and aligned with the intrinsic structure of real-world data, while also introducing an extrapolative mechanism that enhances robustness against out-of-distribution phenomena. CIEGAD addresses this requirement by employing a generation approach that combines interpolation and extrapolation based on the geometric structure within each cluster, effectively complementing semantically uncovered regions both inside and outside the real-world data distribution. The overall process is illustrated in Fig. 3.

For each class-label $y$ and its corresponding cluster $k$, given the embedding set $X_{y,k}$, interpolation-based generation is first performed. Specifically, the outer-edge representative point $x_{y,k,c}$ is selected as the sample with the largest $k$-nearest-neighbor radius (with $k=8$) under cosine distance. Next, the inner set $I_{in}$ is constructed by selecting the 10 samples with the smallest Euclidean distance to the $x_{y,k,c}$, and the outer set $O_{in}$ by selecting the 10 samples with the largest distances. The text corresponding to these samples is included in the prompt provided to the LLM for interpolation. This allows the LLM to treat $O_{in}$ as the outer boundary of the target distribution and to generate samples that remain close to $I_{in}$, effectively

**Algorithm 1**: Hierarchical Frequency–Geometric Allocation for Data Imbalance

Input: the number of samples $N$, class-label $y$, per-class cluster embedding sets $\{X_{y,k}\}$ obtained by $k$-means, centroids $\{\bar{z}_{y,k}\}$, augmentation ratio $\rho$, weights $(\alpha, \beta, \gamma)$
Output: class-level allocations $\{A_y\}$ and per-cluster allocations $\{A_{y,k}\}$.

1: Define the target number of accepted augmented samples:
2: $\quad T = \lfloor \rho \cdot N \rfloor$
3: For each class $y$, count the number of samples $N_y$
4: Compute the inverse-frequency weight for each class to favor minority classes:
5: $\quad w_y = \frac{1}{N_y + \varepsilon}$
6: Calculate the class-level allocation by normalizing these weights so that the total allocation equals $T$:
7: $\quad A_y = \text{round}\left(T \cdot \frac{w_y}{\sum_{y'} w_{y'}}\right)$
8: For each class $y$, distribute $A_y$ across clusters $k$ by computing the cluster-level priority score $v_{y,k}$:
9: $\quad s_{\text{size}} = \frac{1}{N_{y,k} + 10^{-6}}$
10: $\quad s_{\text{sep}} = \min_{k' \neq k} \| \bar{z}_{y,k} - \bar{z}_{y,k'} \|_2$
11: $\quad s_{\text{spar}} = \mathbb{E}_{i \in X_{y,k}}\left[\sqrt{2(1 - \cos(x_{y,k,i}, \bar{z}_{y,k}))}\right]$
12: Compute the overall cluster priority score as the linear combination:
13: $\quad v_{y,k} = \alpha s_{\text{size}} + \beta s_{\text{sep}} + \gamma s_{\text{spar}}$
14: Normalize the priority scores within each class and assign cluster-level allocations:
15: $\quad A_{y,k} = \text{round}\left(A_y \cdot \frac{v_{y,k}}{\sum_{k'} v_{y,k'}}\right)$
16: Return class allocations $\{A_y\}$ and cluster allocations $\{A_{y,k}\}$

suppressing excessive outward expansion. For each set, we compute the centroid vectors $\bar{z}_{y,k}^{I_{in}}$ and $\bar{z}_{y,k}^{O_{in}}$. A newly generated sample embedding $x_{y,k}^{new}$ is accepted as an interpolative sample if it satisfies the following geometric constraint:

$$\left(x_{y,k}^{new} - \bar{z}_{y,k}^{I_{in}}\right)^\top \left(\bar{z}_{y,k}^{O_{in}} - \bar{z}_{y,k}^{I_{in}}\right) \leq 0 \quad (6)$$

This constraint ensures that new samples are located between the inner and outer sets, effectively filling semantically uncovered regions within the existing distribution and achieving smooth, geometry-aware interpolation.

Next, for extrapolative generation, we define the outer set $O_{ex}$ as the 10 samples with the largest distance from the cluster centroid $\bar{z}_{y,k}$, and the inner set $I_{ex}$ as 10 samples whose distances fall within the quantile range [0.70, 0.85] of the distance distribution from the centroid. These sets are incorporated into the LLM prompt for extrapolation. Specifically, $O_{ex}$ is regarded as the innermost boundary of the desired expansion region, while $I_{ex}$ serves as a reference for the opposite, undesired direction, guiding the LLM to generate data outward beyond the current distribution boundary. The centroid vectors $\bar{z}_{y,k}^{O_{ex}}$ and $\bar{z}_{y,k}^{I_{ex}}$ are then computed, and a newly generated embedding $x_{y,k}^{new}$ is accepted as an extrapolative sample if it satisfies the following constraint:

$$\left(x_{y,k}^{new} - \bar{z}_{y,k}^{O_{ex}}\right)^\top \left(\bar{z}_{y,k}^{O_{ex}} - \bar{z}_{y,k}^{I_{ex}}\right) \geq \gamma \cdot |\bar{z}_{y,k}^{O_{ex}} - \bar{z}_{y,k}^{I_{ex}}|_2 \quad (7)$$

where $\gamma = 0.03$ defines the minimum outward displacement threshold. This constraint ensures that new samples are generated beyond the current periphery, preventing them from drifting back toward the cluster center, thereby expanding the boundary of the real-world data distribution. The complete algorithmic pipeline is illustrated in Algorithm 2.

By jointly applying interpolation and extrapolation, CIEGAD effectively controls generation direction and achieves robust, geometry-aware data augmentation that enhances both in-distribution and out-of-distribution generalization.

*E. Data Quality Control for the Real-World Applications*

During both interpolative and extrapolative data generation, the LLM produces 10 new samples per generation cycle for each mode to ensure efficiency. Each prompt explicitly instructs the LLM that all 10 samples must represent distinct contexts and additionally requires the model to generate a brief rationale for each output, thereby maintaining semantic diversity. As a result, a total of 20 new samples is obtained per cycle. To ensure that these generated samples meet the quality standards necessary for real-world applicability, CIEGAD applies two geometric constraint filters, (formulas (6) and (7)), two similarity filters, and an LLM-as-a-Judge–based semantic consistency evaluation.

First, two similarity filters are applied to eliminate redundant or overly similar outputs:
- Intra-batch similarity filtering: To avoid duplicative content among generated samples within the same batch, cosine similarity is computed between each pair of generated texts. Samples with similarity values exceeding a threshold $\theta = 0.85$ are considered semantically redundant and discarded.
- Prompt-overlap filtering: To prevent repetition of content presented in the input prompt (i.e., samples from $I$ and $O$), cosine similarity between generated samples and the prompt references is computed, and samples with similarity exceeding $\theta = 0.9$ are excluded.

Through these filters, only generation outputs that effectively fill semantically uncovered regions are retained for further evaluation.

**Algorithm 2:** Interpolative and Extrapolative Data Generation for Improving Robustness

Input: cluster embedding set $X_{y,k}$, cluster centroid $\bar{z}_{y,k}$, outward margin $\gamma$, LLM, cluster domain profile, per-cluster quota $A_{y,k}$

Output: accepted generated data for cluster $k$ in class $y$

1: Find $x_{y,k,c} \in X_{y,k}$ with the largest $k$-nearest-neighbor radius under cosine distance
2: Form the inner set $I_{in}$ from the $x_{y,k,i} \in X_{y,k}$ with the smallest Euclidean distance to $x_{y,k,c}$
3: Form the outer set $O_{in}$ from the $x_{y,k,i} \in X_{y,k}$ with the largest Euclidean distance to $x_{y,k,c}$
4: Compute set centroids $\bar{z}_{y,k}^{I_{in}}$ and $\bar{z}_{y,k}^{O_{in}}$
5: Build an interpolation prompt including the cluster's domain profile and text from $I_{in}$ and $O_{in}$
6: Generate a new sample with the LLM from the prompt, then encode it to obtain $x_{y,k}^{new}$
7: Accept a candidate as interpolative if
8: $\left(x_{y,k}^{new} - \bar{z}_{y,k}^{I_{in}}\right)^\top \left(\bar{z}_{y,k}^{O_{in}} - \bar{z}_{y,k}^{I_{in}}\right) \leq 0$
9: Form the inner set $O_{ex}$ from the $x_{y,k,i} \in X_{y,k}$ with the largest Euclidean distance to $\bar{z}_{y,k}$
10: Form the outer set $I_{ex}$ from the $x_{y,k,i} \in X_{y,k}$ whose distances to $\bar{z}_{y,k}$ fall in the quantile range [0.70, 0.85]
11: Compute set centroids $\bar{z}_{y,k}^{O_{ex}}$ and $\bar{z}_{y,k}^{I_{ex}}$
12: Build an extrapolation prompt including the cluster's domain profile and text from $O_{ex}$ and $I_{ex}$
13: Generate a new sample with the LLM from the prompt, then encode it to obtain $x_{y,k}^{new}$
14: Accept a candidate as extrapolative if
15: $\left(x_{y,k}^{new} - \bar{z}_{y,k}^{O_{ex}}\right)^\top \left(\bar{z}_{y,k}^{O_{ex}} - \bar{z}_{y,k}^{I_{ex}}\right) \geq \gamma \cdot |\bar{z}_{y,k}^{O_{ex}} - \bar{z}_{y,k}^{I_{ex}}|_2$
16: Add all accepted interpolative/extrapolative samples to $X_{y,k}$
17: Repeat Steps 1–16 until the total number of accepted samples reaches the quota $A_{y,k}$

Next, we apply the LLM-as-a-Judge assessment, which evaluates the semantic and stylistic quality of generated samples based on both the cluster's domain profile and the generated text. Each sample is scored on a five-point Likert scale across the following five dimensions:
1. Emotion consistency
2. Style alignment
3. Lexical/topic coherence
4. Diversity
5. Reason validity (interpolation/extrapolation)

A sample is accepted if its average score is 3 or higher, prioritizing diverse candidates that maintain cluster alignment rather than enforcing strict stylistic uniformity.

After applying the geometric constraint filters, similarity filters, and the LLM-as-a-Judge evaluation sequentially, the remaining high-quality samples are accepted as augmented data. These accepted samples are then added to the corresponding cluster's dataset $X_{y,k}$, updating its composition. After each update, new inner and outer sets are selected, and both interpolative and extrapolative generations are repeated until the allocated number of samples $A_{y,k}$ is satisfied. This iterative process allows CIEGAD to preserve the advantages of conventional data augmentation while enhancing robustness against outliers and out-of-distribution phenomena frequently encountered in real-world applications.

## IV. Experiment Design

### A. Dataset

To validate the effectiveness of the proposed framework CIEGAD, we employed four representative real-world datasets targeting sentiment polarity classification and emotion recognition tasks:

- IMDb [32]: A binary sentiment polarity classification dataset consisting of movie reviews. The texts are relatively long and descriptive, labeled as *positive* or *negative*.
- Yelp [33]: A large-scale user review dataset for sentiment analysis. It contains short and colloquial expressions, also labeled as *positive* or *negative*.
- Emo-6class [34]: A Twitter-based emotion classification dataset containing six emotion labels. It exhibits moderate class imbalance.
- Emo-13class [35]: A fine-grained emotion recognition dataset with 13 emotion labels. It shows substantial overlap between classes and follows a long-tail distribution, resulting in severe class imbalance.

These datasets collectively cover a broad spectrum of linguistic and distributional characteristics, from binary sentiment to multi-emotion classification, enabling a comprehensive evaluation of the generalization capability of CIEGAD. In this study, 10,000 samples were randomly selected from IMDb and Yelp, and 15,000 samples were drawn from Emo-6class and Emo-13class while preserving their original label distributions.

### B. Baseline and Evaluation Methods

To assess the performance of CIEGAD, we compared it with five baseline methods, ranging from simple LLM-based augmentation to recent advanced generation approaches:

- SynReLLM: The LLM is instructed to replace one to three words with synonyms while preserving meaning, performing lightweight lexical-level augmentation that maintains semantic and label consistency.
- PerReLLM: Each input sentence is rewritten from the perspective of a predefined persona description, generating expressions as if spoken by that persona. This approach controls social tone and emotional expression during augmentation.
- Self-Augmentation [12]: A self-training approach in which the LLM reuses its own generated outputs as additional training data. We reproduced the same self-generation and retraining procedure within our experimental framework for fair comparison.
- Mini-DA [11]: A few-shot LLM-based augmentation method that generates one to three new sentences per sample using a interpolative strategy. It represents a lightweight and locally focused augmentation approach.
- LM-DA [15]: A conditional generation framework that utilizes LLMs to generate samples according to the class-label distribution. It focuses on alleviating class imbalance through label-aware data generation.

All baselines were implemented within the same evaluation pipeline and used identical downstream classification models for a fair comparison.

To comprehensively evaluate the qualitative characteristics of generated data from each method, we employed the following three metrics:

- Outward Expansion Ratio (OER): A geometric measure indicating the degree to which generated samples extend beyond the boundary of the original data distribution. Excessively large values suggest distributional drift, while moderate values reflect effective coverage of semantically uncovered regions.
- Semantic Novelty Index (SNI): A metric quantifying the semantic uniqueness of generated samples, capturing new lexical and contextual diversity while maintaining label consistency. Higher values indicate meaningful semantic diversity.
- Fréchet Embedding Distance (FED): A statistical distance measure between the embedding distributions of original and generated data. Smaller values indicate higher distributional alignment and domain consistency.

These three indicators jointly evaluate domain alignment (FED), semantic diversity (SNI), and distributional adequacy (OER), providing a balanced and multidimensional assessment of each augmentation strategy.

For all generation processes, we used GPT-4.1-mini [36] via the OpenAI API as the data generator. For the LLM-as-a-Judge evaluation, the more capable GPT-4.1 model was employed through the same API. Both models were used with their default parameters. Additionally, all text embeddings were obtained using the Sentence Transformer [32] without any additional fine-tuning. Under these configurations, we conducted a comprehensive evaluation of CIEGAD across all datasets and baselines.

## V. Experiment and Analysis

### A. Evaluation on Data Augmentation for Downstream Tasks

We first generated augmented data using all baseline methods and the proposed CIEGAD framework for the IMDb, Yelp, Emo-6class, and Emo-13class datasets, and computed the three evaluation metrics, OER, SNI, and FED. The results are presented in Fig. 4. CIEGAD achieved the highest OER values across all datasets, indicating its capability to effectively fill the semantic gaps in the original distributions. Moreover, CIEGAD recorded the highest SNI scores in all cases, demonstrating its ability to enhance semantic diversity while maintaining class-label consistency. In addition, its FED values remained moderate compared with other methods, implying that excessive deviation from the real-world data distributions was successfully suppressed, and that the augmented data maintained strong domain alignment. These findings collectively confirm that CIEGAD is the only method that simultaneously satisfies the three criteria, high diversity, distributional consistency, and appropriate expansion, effectively complementing semantically uncovered regions both within and beyond real-world data distributions.

Next, Fig. 5 presents the t-SNE visualization results of the embeddings for real-world and augmented data in the Emo-13class dataset. The blue dots represent real-world data, while the orange dots indicate augmented data. In CIEGAD, the orange points are moderately expanded along the semantically uncovered regions, smoothly filling intra-class gaps while suppressing excessive outward expansion. As a result, the overall distribution is continuously and structurally extended, showing strong alignment between real and augmented data in high-dimensional space. In contrast, SynReLLM and PerReLLM mainly perform augmentation along the existing data manifold, with limited extrapolative effects. Self-Augment and LM-DA fail to sufficiently cover semantically uncovered regions, while Mini-DA achieves relatively better performance but still generates redundant samples close to existing data. These visualizations directly demonstrate the effectiveness of CIEGAD's direction control mechanism and geometric filtering, showing that data generation is faithfully regulated along the intrinsic cluster structure.

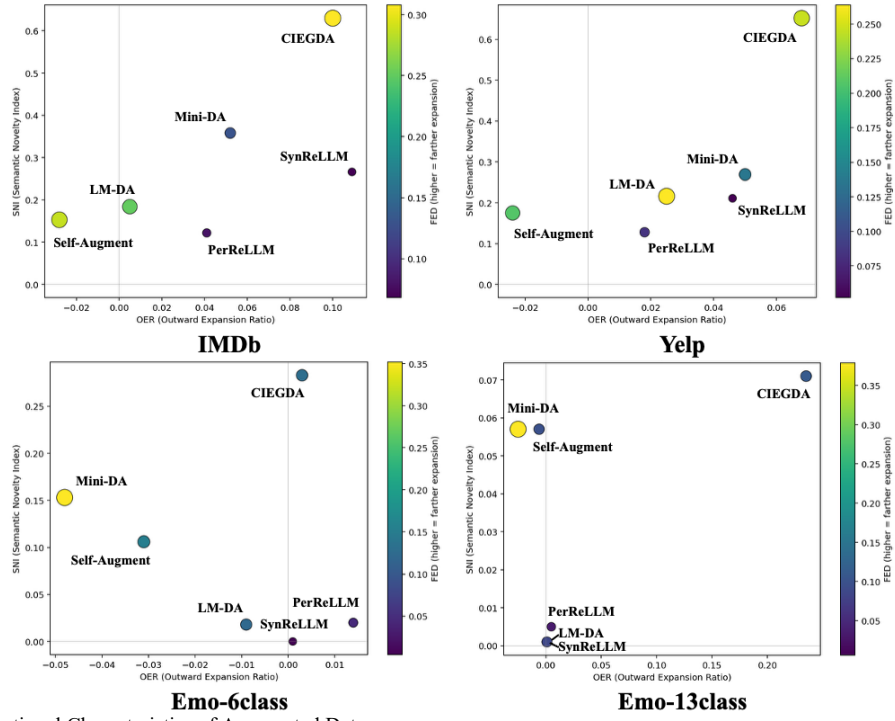

Fig. 4. Evaluation of Distributional Characteristics of Augmented Data.

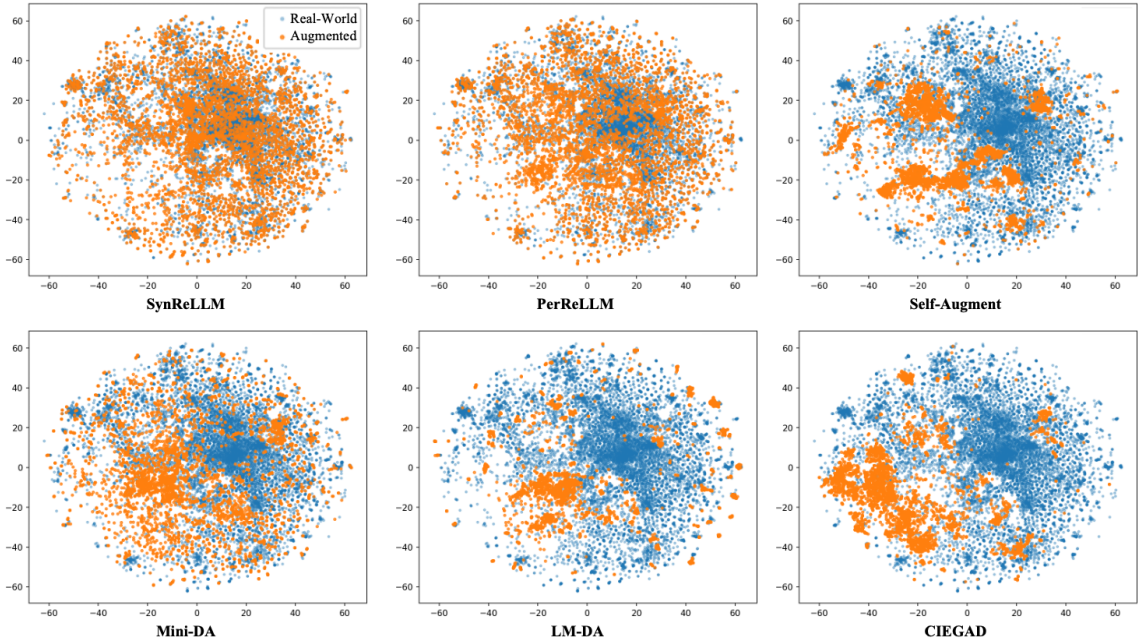

Fig. 5. Evaluation of Distributional Alignment via t-SNE Visualization.

In addition, we trained BERT models [37] using the combination of real-world and augmented data for each dataset and compared the performance against a baseline model trained only on real-world data (Golden). The evaluation metrics included accuracy, precision, recall, and F1. For IMDb and Yelp, 2,000 test samples were used, while 3,000 were used for Emo-6class and Emo-13class. The results are summarized in Table I. CIEGAD consistently outperformed all other methods on the Emo-6class and Emo-13class datasets, with especially notable improvements in F1 and recall. These datasets have multiple class-labels with long-tailed distributions, and thus the hierarchical frequency–geometric allocation in CIEGAD effectively reinforced minority classes and peripheral clusters, contributing to improved learning robustness. For IMDb and Yelp, which are binary classification tasks, the performance gap among methods was small because these datasets are relatively saturated. Nevertheless, CIEGAD maintained stable and competitive performance without degradation, preventing over-expansion while preserving top-level accuracy. Overall, these results demonstrate that CIEGAD is particularly effective for data augmentation under multi-class and long-tailed settings, improving model generalization by jointly achieving distributional alignment and semantic diversity.

TABLE I
EVALUATION OF DOWNSTREAM TASK PERFORMANCE WITH AUGMENTED DATA

| Dataset | IMDb | | | | Yelp | | | | Emo-6class | | | | Emo-13class | | | |
|---|---|---|---|---|---|---|---|---|---|---|---|---|---|---|---|---|
| Framework | Accu | Pre | Rec | F1 | Acc | Pre | Rec | F1 | Acc | Pre | Rec | F1 | Acc | Pre | Rec | F1 |
| Golden | 0.872 | 0.872 | 0.872 | 0.871 | 0.941 | 0.941 | 0.941 | 0.940 | 0.876 | 0.852 | 0.758 | 0.785 | 0.549 | 0.635 | 0.427 | 0.433 |
| SynReLLM | 0.877 | 0.877 | 0.877 | 0.877 | 0.942 | 0.942 | 0.942 | 0.941 | 0.899 | 0.872 | 0.852 | 0.861 | 0.560 | 0.645 | 0.460 | 0.470 |
| PerReLLM | 0.872 | 0.872 | 0.872 | 0.872 | 0.948 | 0.948 | 0.948 | 0.948 | 0.906 | 0.870 | 0.873 | 0.871 | 0.576 | 0.621 | 0.500 | 0.518 |
| Self-Augment | 0.885 | 0.885 | 0.885 | 0.884 | 0.947 | 0.947 | 0.947 | 0.946 | 0.884 | 0.859 | 0.810 | 0.829 | 0.566 | 0.615 | 0.494 | 0.510 |
| Mini-DA | 0.881 | 0.881 | 0.881 | 0.880 | 0.947 | 0.947 | 0.947 | 0.946 | 0.886 | 0.857 | 0.792 | 0.815 | 0.549 | 0.626 | 0.448 | 0.450 |
| LM-DA | 0.881 | 0.881 | 0.881 | 0.880 | 0.948 | 0.948 | 0.948 | 0.948 | 0.901 | 0.856 | 0.880 | 0.867 | 0.569 | 0.572 | 0.532 | 0.544 |
| CIEGAD | 0.882 | 0.882 | 0.882 | 0.881 | 0.945 | 0.945 | 0.945 | 0.944 | 0.906 | 0.859 | 0.893 | 0.873 | 0.582 | 0.576 | 0.553 | 0.560 |

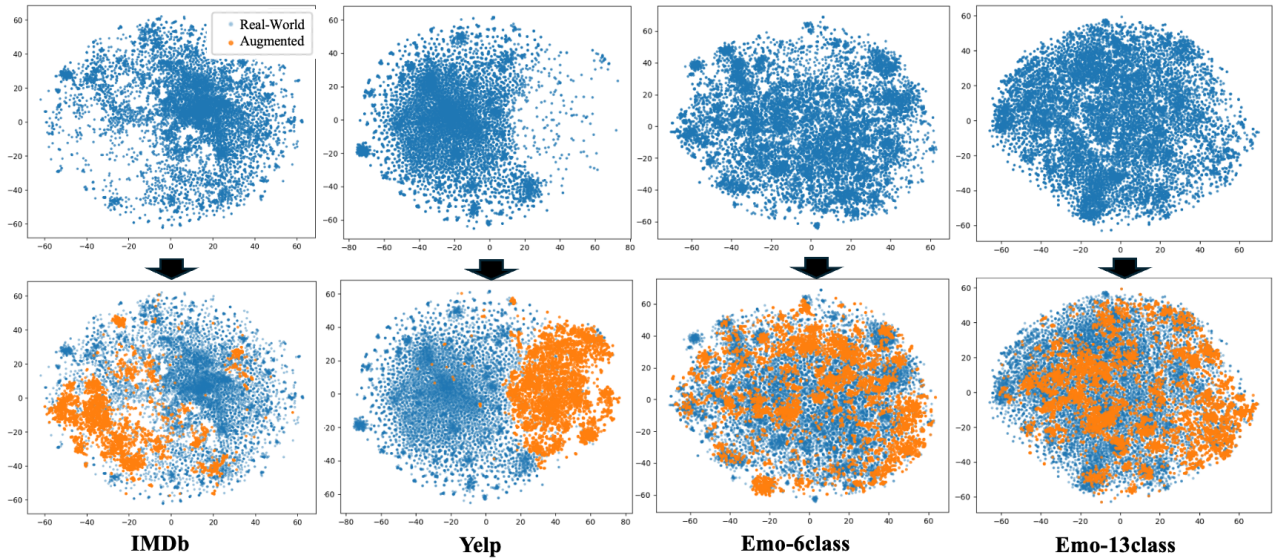

Fig. 6. Evaluation of Augmented Data Placement across Datasets.

### B. Evaluation on Augmented Data Quality

Fig. 6 visualizes the distributional structure of augmented data generated by CIEGAD for IMDb, Yelp, Emo-6class, and Emo-13class. As before, blue points denote real-world data, and orange points represent augmented data. Across all datasets, the augmented data from CIEGAD naturally populate the semantically uncovered regions of the real-world distributions, effectively filling existing gaps. For IMDb and Yelp, the augmented samples clearly extend the data manifold into underrepresented peripheral regions, while for Emo-6class and Emo-13class, both intra-distribution and out-of-distribution uncovered areas are systematically complemented. Notably, few samples are generated in already dense regions, showing that CIEGAD performs augmentation efficiently by adapting to the inherent density of the data distribution. This indicates that CIEGAD's interpolative and extrapolative mechanism flexibly adjusts to the geometric characteristics of the dataset. Consequently, CIEGAD does not simply push data outward but instead dynamically adjusts the generation direction based on the geometric structure of the real-world data, achieving distributionally aligned augmentation that effectively fills semantically uncovered regions within and beyond the data manifold.

To further examine the effect of the hierarchical frequency–geometric allocation, we compared class-label distributions before and after augmentation, as shown in Fig. 7. The left panel corresponds to Emo-6class, and the right to Emo-13class, where blue bars represent the proportion of real-world data and green bars denote the proportion after augmentation. In the original datasets, class imbalance was severe, with samples concentrated in a few dominant labels such as *joy*, *sadness*, and *happiness*, while others were extremely underrepresented. Through inverse-frequency weighting and geometric–density-based allocation, CIEGAD automatically allocated more generation resources to minority and peripheral clusters. As a result, the post-augmentation distributions became significantly more balanced, simultaneously mitigating class-label imbalance and equalizing inter-cluster density. In particular, in Emo-13class, low-frequency labels such as *desire*, *guilt*, and *sarcasm* were substantially reinforced, demonstrating the effectiveness of CIEGAD's hierarchical frequency–geometric allocation for long-tailed, multi-class scenarios. This finding supports that CIEGAD performs not a mere quantitative rebalancing but a geometry-aware redistribution reflecting the underlying data structure.

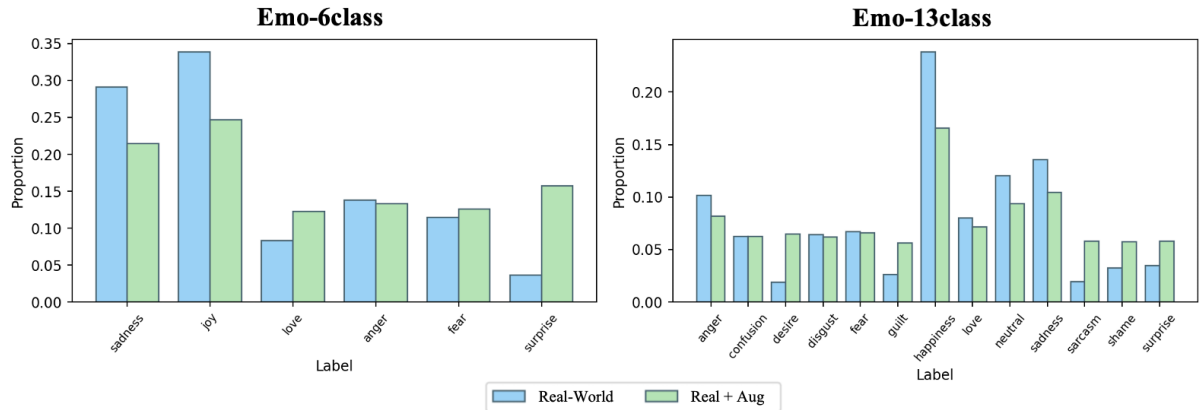

Fig. 7. Evaluation of Class-Label Rebalancing via HFGA.

TABLE II
EVALUATION OF QUALITATIVE CHARACTERISTICS OF GENERATED TEXT

| Label | Type | Example Sentences (excerpt) | Qualitative Characteristics |
|---|---|---|---|
| Sarcasm | Real-World | "Wow, error messages that explain nothing, clarity champion!" "Yay, CR sensors reject tickets, entry challenge!" | High stylistic coherence with sarcastic tone and humor; limited lexical diversity but strong pragmatic consistency. |
| Sarcasm | Augmented | "Perfect, the update broke everything again, innovation at its best!" "Wonderful, more bugs to keep my weekend productive!" | Preserves sarcastic expressions while enriching lexical variation and pragmatic intensity, maintaining contextual fluency. |
| Desire | Real-World | "I want to visit Auckland next fall." "I crave a new bhaji dish." | Expresses simple motivational intent with direct wording and moderate emotional intensity. |
| Desire | Augmented | "I'm longing to learn meditation this year." "I wish I could rebuild my pantry from scratch." | Expands lexical and contextual range while preserving the affective tone of aspiration; introduces nuanced emotional depth. |

Finally, to assess the qualitative characteristics of generated text, Table II presents representative examples from the *sarcasm* and *desire* classes in the Emo-13class dataset, comparing real-world and augmented data. The results show that CIEGAD expands lexical and contextual diversity while preserving each class's emotional and stylistic characteristics. In the *sarcasm* class, the generated texts maintain sarcastic tone and syntactic patterns while broadening semantic variation, whereas in the *desire* class, expressions with different emotional intensities and target objects are generated. This is attributed to the use of cluster-level domain profiles as conditioning information in generation, enabling stylistically aligned yet semantically diverse augmentation that simple paraphrasing cannot achieve. Overall, CIEGAD consistently demonstrates high-quality characteristics across distributional, statistical, and semantic dimensions of the augmented data.

### C. Evaluation on Prediction Robustness

To evaluate how effectively the augmented data generated by CIEGAD improves the robustness of downstream task models, we conducted experiments on the Emo-13class dataset by gradually increasing the size of the test set. Specifically, the number of test samples was expanded stepwise to 3,000, 5,000, 10,000, 30,000, and 50,000. For each stage, we calculated accuracy, precision, recall, and F1, and compared the results with the baseline model trained only on real-world data (Golden). The results are summarized in Table III, where the values in parentheses indicate the change relative to the case with 3,000 test samples. For the Golden model, all metrics gradually declined as the test size increased. When the test set reached 50,000 samples, accuracy decreased by -0.012 and F1 by -0.008. The decline in recall was particularly pronounced, suggesting that as the test scale grew, the model's generalization ability toward minority classes deteriorated.

In contrast, CIEGAD maintained nearly stable performance, with accuracy changing only slightly from 0.582 to 0.579 (-0.003) and F1 from 0.560 to 0.555 (-0.005). Precision and recall also remained stable within a narrow range of 0.000 to -0.006. Even when the test scale increased more than tenfold, performance degradation was minimal. This stability can be attributed to CIEGAD's ability to maintain strong distributional alignment while effectively complementing semantically uncovered regions both within and beyond the data distribution, resulting in consistent generalization to unseen data.

These findings demonstrate that CIEGAD exhibits extremely low performance decay even as test scale increases, achieving both high robustness and generalization capability. In particular, for emotion classification tasks with long-tailed label structures, CIEGAD maintains prediction accuracy and stability under large-scale conditions, indicating its strong applicability to real-world applications.

TABLE III
EVALUATION OF PREDICTION ROBUSTNESS UNDER INCREASING TEST SCALE

| Framework | Test Data | Accuracy | Precision | Recall | F1 |
|---|---|---|---|---|---|
| Golden | 3000 | 0.549 | 0.635 | 0.427 | 0.433 |
|  | 5000 | 0.536 (-0.013) | 0.625(-0.010) | 0.418(-0.009) | 0.427(-0.006) |
|  | 10000 | 0.533(-0.016) | 0.633(-0.002) | 0.415(-0.012) | 0.424(-0.009) |
|  | 30000 | 0.535(-0.014) | 0.637(0.002) | 0.413(-0.014) | 0.424(-0.009) |
|  | 50000 | 0.537(-0.012) | 0.643(0.008) | 0.414(-0.013) | 0.425(-0.008) |
| CIEGAD | 3000 | 0.582 | 0.576 | 0.553 | 0.560 |
|  | 5000 | 0.576(-0.006) | 0.575(-0.001) | 0.551(-0.002) | 0.556(-0.004) |
|  | 10000 | 0.576(-0.006) | 0.573(-0.003) | 0.553(0.000) | 0.553(-0.007) |
|  | 30000 | 0.578(-0.004) | 0.572(-0.004) | 0.547(-0.006) | 0.554(-0.006) |
|  | 50000 | 0.579(-0.003) | 0.573(-0.003) | 0.548(-0.005) | 0.555(-0.005) |

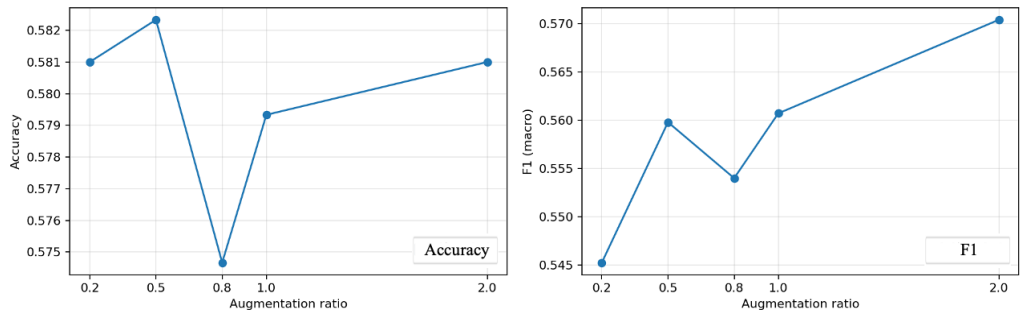

Fig. 8. Evaluation of Augmented Data Expansion Efficiency.

*D. Evaluation on Data Expansion Efficiency*

To assess the data expansion efficiency of CIEGAD, we analyzed how varying the augmentation ratio $\rho$ affects model performance. The augmentation ratio $\rho$ represents the proportion of augmented data relative to the number of real-world data N, and the total accepted number of augmented data is defined as $T = \lfloor \rho N \rfloor$. In this experiment, we set $\rho \in \{0.2, 0.5, 0.8, 1.0, 2.0\}$ and evaluated the changes in accuracy and F1 on the Emo-13class dataset. The results are illustrated in Fig. 8.

The results show that accuracy peaked at $\rho = 0.5$ and slightly decreased or plateaued beyond that point. This trend suggests that during the initial phase of data augmentation, the representation space becomes quickly enriched, whereas excessive augmentation introduces redundancy and reduces learning efficiency due to the proliferation of similar samples within clusters. Therefore, $\rho \approx 0.5$ represents the most cost-effective augmentation condition. On the other hand, the F1 showed a consistent upward trend as $\rho$ increased, reaching its maximum at $\rho = 2.0$. This improvement indicates that increasing the amount of augmented data enhances learning opportunities for minority classes, thereby improving overall class balance and recall. Importantly, since CIEGAD suppresses generation noise through its geometric constraints and cluster-conditioned synthesis, the model maintains stability even with large-scale augmentation, and the increased quantity directly contributes to better F1 performance.

In summary, CIEGAD achieves substantial performance gains even with a moderate augmentation ratio ($\rho \approx 0.5$), while allowing further improvement in F1 through larger-scale augmentation. This demonstrates that CIEGAD effectively balances efficiency and scalability, enabling flexible augmentation strategies that can adapt to various resource constraints and target task requirements.

*E. Ablation Study*

To investigate how each component of the proposed CIEGAD framework contributes to the generated data distribution and overall data quality, we conducted an ablation study. Four variant models were designed by removing specific modules, and data augmentation was performed on the Emo-13class dataset to quantitatively assess the contribution of each component. The four models are summarized as follows:

- w/o Profile: Removes the cluster-level domain profile and generates samples conditioned only on class-labels. This configuration evaluates the effect of ignoring cluster-specific domain structures on distributional alignment.
- w/o Extrapolation: Performs only interpolation-based generation, restricting augmentation to within the central regions of the distribution. This setup assesses the impact of strengthening intra-distribution density on overall performance.
- w/o Interpolation: Performs only extrapolation-based generation, limiting augmentation to outward expansion along the periphery of the distribution. This configuration analyzes how expansion toward the outer regions affects data quality and stability.
- w/o Geometry: Disables the geometric constraint filter, effectively removing the directional control mechanism. This variant evaluates the contribution of geometric constraints to stabilizing generation direction and preventing excessive expansion.

Table IV presents the results of these variants across three evaluation metrics: OER, SNI, and FED. The values in parentheses indicate the difference relative to the full CIEGAD framework.

TABLE IV
EVALUATION OF COMPONENT CONTRIBUTIONS IN ABLATIONS STUDY

| Framework | OER | SNI | FED |
|---|---|---|---|
| CIEGAD | 0.235 | 0.071 | 0.112 |
| w/o Profile | 0.393(+0.158) | 0.267(+0.196) | 0.186(+0.074) |
| w/o Extrapolation | 0.205(-0.030) | 0.051(-0.020) | 0.176(+0.064) |
| w/o Interpolation | 0.281(+0.046) | 0.126(+0.055) | 0.166(+0.054) |
| w/o Geometry | 0.245(+0.010) | 0.089(+0.018) | 0.144(+0.032) |

The results show that w/o Profile recorded a substantial increase in OER, indicating excessive expansion of the generated distribution. Both SNI and FED also deteriorated significantly, confirming that without the domain profile, the generated data failed to preserve local structure and lost domain alignment. Thus, cluster-conditioned domain control serves as a core mechanism of CIEGAD. Next, the results for w/o Extrapolation and w/o Interpolation highlight the necessity of combining both interpolation and extrapolation. In w/o Extrapolation, both OER and SNI decreased, and FED worsened by +0.064, suggesting insufficient novelty and degraded quality. Conversely, w/o Interpolation resulted in increased OER, SNI, and FED, indicating overexpansion and instability in the generated distribution. This contrasting behavior confirms that CIEGAD's balanced design, complementing semantically uncovered regions both within and outside the data manifold, is critical for achieving an equilibrium between diversity and stability. Finally, w/o Geometry degraded all three metrics, showing that the absence of geometric constraints caused the generated samples to drift away from the original distribution without proper directional regulation. This demonstrates the importance of the geometric constraint filter in stabilizing generation and preventing deviation from the data manifold.

Overall, these results indicate that the components of CIEGAD function in a complementary manner. In particular, the combination of domain-profile conditioning, joint interpolation and extrapolation, and geometric constraint filtering enables simultaneous achievement of distributional alignment, semantic diversity, and high data quality. Consequently, CIEGAD is validated as a robust and structurally coherent data generation framework that systematically covers semantically uncovered regions while suppressing overexpansion.

*F. Evaluation on Transferability*

To assess the transferability of the proposed CIEGAD framework, we examined how its behavior changes when replacing the LLM used for data generation and evaluation. While GPT models were primarily used in the main experiments, we also evaluated two alternative LLM families, Gemini [38] and Claude [39], to verify cross-model adaptability. The configurations for each case are as follows.
- Gemini: Data generation with Gemini-2.5-Flash-Lite and LLM-as-a-Judge evaluation with Gemini-2.5-Flash.
- Claude: Data generation with Claude-3-Haiku and LLM-as-a-Judge evaluation with Claude-3.5-Haiku.

All models were used with their default parameters. The results for IMDb, Yelp, Emo-6class, and Emo-13class are summarized in Table V, where the same three evaluation metrics were computed for comparison.

Overall, GPT exhibited the lowest OER and SNI values among the models, indicating that it strongly suppresses excessive expansion of the data distribution. Meanwhile, its FED values were consistently lower than those of other models, demonstrating a strong emphasis on maintaining distributional alignment. These tendencies suggest that GPT is well-suited for balanced augmentation tasks in practical settings, where preserving label integrity while ensuring moderate diversity is critical. In contrast, Gemini produced the highest OER and SNI scores, revealing a strong ability to complement semantically uncovered regions. Although its FED values tended to be higher than those of GPT, the results remained stable overall, indicating that Gemini performs extrapolative generation effectively without severely compromising domain consistency. This characteristic makes Gemini particularly suitable for exploratory augmentation scenarios that prioritize expanding coverage into unseen distributions. Finally, Claude displayed results generally similar to GPT, but with notably smaller OER values for the Emo-13class dataset. This suggests that Claude is less capable of outward expansion and instead excels at stable, conservative augmentation within the distribution. Such behavior implies that Claude is best suited for domains requiring high-precision and reliable data supplementation rather than extrapolative generation.

In summary, these findings confirm that CIEGAD is a flexible framework capable of supporting a wide spectrum of augmentation strategies, from controlled to exploratory, depending on the characteristics of the underlying LLM. By selecting the appropriate model according to task objectives, users can tailor CIEGAD to achieve either conservative domain-consistent augmentation or broader exploratory data expansion.

## VI. DISCUSSION

*A. Key Findings*

The results of this study demonstrate that the proposed CIEGAD framework serves as a general-purpose data augmentation approach capable of complementing semantically uncovered regions both within and beyond real-world data distributions while maintaining strong distributional alignment. Specifically, the introduction of cluster-conditioned domain profiles proved crucial for locally controlling generation direction while preserving cluster-specific lexical, stylistic, and emotional tendencies. This mechanism effectively retained local domain structures that are often lost in conventional class-label-only generation, ensuring that augmented data were coherently positioned within the real-world data distribution. Furthermore, by jointly employing interpolation and extrapolation generation, CIEGAD successfully complemented semantically uncovered regions across both central and peripheral distributional areas, achieving simultaneous improvement in OER and SNI while maintaining stable FED values.

The hierarchical frequency–geometric allocation further contributed to alleviating data imbalance by integrating class-label frequency and geometric cluster characteristics, prioritizing the augmentation of minority classes and peripheral clusters. Consequently, in downstream emotion classification tasks, particularly in the long-tailed Emo-13class dataset, CIEGAD achieved substantial improvements in F1 and recall. These findings confirm that CIEGAD strategically mitigates structural imbalance in class-label distributions and enhances generalization performance in a sustained manner. Moreover, the three-layer quality control mechanism combining geometric constraint filtering, similarity filtering and LLM-as-a-Judge effectively prevented excessive extrapolation while ensuring linguistic consistency and stylistic coherence. The ablation study corroborated that the removal of these components substantially degraded OER and FED, highlighting that the integration of CIEGAD's components directly contributes to stability and robustness in generation.

TABLE V
EVALUATION OF TRANSFERABILITY ACROSS LLM FAMILIES

| Dataset | IMDb | | | Yelp | | | Emo-6class | | | Emo-13class | | |
|---|---|---|---|---|---|---|---|---|---|---|---|---|
| Model | OER | SNI | FED | OER | SNI | FED | OER | SNI | FED | OER | SNI | FED |
| GPT | 0.100 | 0.630 | 0.308 | 0.068 | 0.652 | 0.246 | 0.003 | 0.283 | 0.129 | 0.235 | 0.071 | 0.112 |
| Gemini | 0.149 | 0.769 | 0.290 | 0.083 | 0.808 | 0.276 | 0.034 | 0.504 | 0.179 | 0.004 | 0.122 | 0.115 |
| Claude | 0.166 | 0.638 | 0.290 | 0.071 | 0.642 | 0.270 | 0.005 | 0.291 | 0.183 | 0.001 | 0.101 | 0.137 |

In addition, experiments with varying augmentation ratios $\rho$ demonstrated that CIEGAD achieves significant performance gains even with small-scale augmentation ($\rho \approx 0.5$), indicating high efficiency under limited generation resources, a practically valuable insight for real-world applications. Transferability experiments further showed that CIEGAD maintains stable performance regardless of the underlying LLM, while leveraging model-specific generation characteristics. This confirms that CIEGAD functions as a flexible and extensible framework capable of supporting a full spectrum of augmentation, from controlled to exploratory, depending on the choice of generation model.

Overall, CIEGAD successfully balances distributional alignment, semantic diversity, and data quality, providing a stable and efficient augmentation strategy that enhances learning performance even under limited or imbalanced data conditions.

*B. Limitations*

Despite its effectiveness, this study has several limitations. First, the scope of evaluation was restricted to datasets in the emotion recognition domain. This choice was deliberate, as emotion recognition focuses primarily on maintaining fixed emotional polarity while introducing semantic diversity, making it relatively tractable for controlled augmentation. However, more complex domains, such as news classification or policy document analysis, require stricter factual consistency and stylistic precision, where simply increasing generation diversity may not yield appropriate augmentation. Future work should therefore explore mechanisms that can dynamically balance generation diversity and fidelity according to task characteristics.

Second, in both interpolation and extrapolation processes, the number of samples in the inner set and outer set was fixed at ten. Although this configuration produced stable results, the validity of this setting has not been fully examined. Increasing the number of samples may allow for finer directional control in generation, but it also introduces computational overhead and the risk of generating redundant data. A systematic analysis of how these parameters influence generation quality and expansion efficiency remains an important direction for future refinement. Third, the geometric constraint filter was applied in a fixed formulation. There remains potential to design more flexible mechanisms by dynamically adjusting the strength of constraints according to local cluster density or similarity. Such adaptive constraints could enable more context-aware augmentation, efficiently covering underrepresented boundary regions while suppressing unnecessary outward expansion.

Additionally, all generations in this study were performed using closed but high-performing LLM APIs. While these APIs ensure high-quality outputs, they present challenges in terms of reproducibility and computational cost. If the CIEGAD framework could be reproduced using open-source or lightweight local models, its practical utility would greatly increase. However, preliminary experiments revealed that smaller open-source models struggle with precise output control, leading to inconsistent and lower-quality generation. Future research should therefore focus on improving controllability and coherence in lightweight models, enabling broader applicability of CIEGAD in resource-constrained settings. Finally, this study focused exclusively on text data. Extending CIEGAD's principles to multimodal data, including images, audio, and physiological signals, represents a promising direction. Incorporating cross-modal interpolation and extrapolation strategies that integrate feature spaces across modalities could facilitate the reconstruction of complex real-world data distributions, leading to the development of a more powerful and versatile augmentation framework.

In summary, while CIEGAD demonstrated remarkable effectiveness in emotion recognition tasks, further advancements are needed to extend its applicability across data domains, model scales, geometric constraints, and modalities. Addressing these challenges will enable CIEGAD to evolve into a universal and practically deployable data augmentation foundation.

VII. CONCLUSION

This study proposed CIEGAD, a cluster-conditioned interpolative and extrapolative framework for geometry-aware and domain-aligned data augmentation that systematically complements semantically uncovered regions within and beyond real-world data distributions. CIEGAD integrates a cluster-conditioned domain profiling mechanism, a hierarchical frequency–geometric allocation strategy for balanced data generation, directional control through joint interpolation and extrapolation, and a multi-layer quality control pipeline combining geometric constraints with LLM-based evaluation.

15Through this integrated design, CIEGAD achieves a high-level balance of distributional alignment, semantic diversity, and data quality, a combination that has been difficult for prior augmentation methods to attain, while consistently enhancing learning performance under limited or imbalanced data conditions.

Experiments on four representative datasets, IMDb, Yelp, Emo-6class, and Emo-13class, demonstrated the effectiveness of the proposed framework. Notably, on the Emo-13class dataset, which contains numerous labels and exhibits a long-tailed structure, CIEGAD achieved the highest F1 and recall among all methods. The framework successfully maximized OER and SNI while stabilizing FED, indicating its ability to moderately control outward expansion and maintain domain consistency. In contrast, on denser datasets such as IMDb and Yelp, the improvement was more limited due to the already saturated data distributions, suggesting that CIEGAD is particularly advantageous in sparse or long-tailed environments.

Future work will include further validation across more diverse datasets, as well as systematic investigation of design parameters such as the number of samples in interpolation and extrapolation and the dynamic adaptation of geometric constraint filters. Moreover, while this study utilized high-quality LLM APIs such as GPT-4.1-mini, developing a lightweight implementation reproducible with open-source models remains an important step toward practical deployment. Finally, extending the principles of CIEGAD beyond text to multimodal data, including images, audio, and physiological signals, could establish it as a general foundation for cross-modal data augmentation. In conclusion, CIEGAD introduces a new paradigm of data augmentation that unifies structural and semantic coherence, offering a scalable and robust framework capable of enhancing learning performance across diverse domains and data environments.

## References

[1] Q. Xie, P. Zhang, B. Yu, and J. Choi, "Semisupervised training of deep generative models for high-dimensional anomaly detection," *IEEE Transactions on Neural Networks and Learning Systems*, vol. 33, no. 6, pp. 2444–2453, June. 2022.

[2] L. Alzubaidi et al., "A survey on deep learning tools dealing with data scarcity: definitions, challenges, solutions, tips, and applications," *Journal of Big Data*, vol. 10, no. 1, pp. 1–82, Apr. 2023.

[3] M. Bayer, M.-A. Kaufhold, and C. Reuter, "A survey on data augmentation for text classification," *ACM Computing Surveys*, vol. 55, no. 7, pp. 1–39, July. 2023.

[4] Y. Li, Y. Zhang, Z. Du, and Z. Guo, "Large language model data augmentation for text-pair classification tasks," in *Proceedings of the 2024 13th International Conference on Computing and Pattern Recognition*, pp. 427–433, Jan. 2025.

[5] G. Kambhatla, C. Shaib, and V. S. Govindarajan, "Measuring lexical diversity of synthetic data generated through fine-grained persona prompting," in *Findings of the Association for Computational Linguistics: EMNLP 2025*, pp. 21024–21033, Nov. 2025.

[6] J. Mai, C. Gao, and J. Bao, "Domain generalization through data augmentation: A survey of methods, applications, and challenges," *Mathematics*, vol. 13, no. 5, p. 824, Feb. 2025.

[7] I. Shumailov, Z. Shumaylov, Y. Zhao, N. Papernot, R. Anderson, and Y. Gal, "AI models collapse when trained on recursively generated data," *Nature*, vol. 631, no. 8022, pp. 755–759, July. 2024.

[8] J. Wei and K. Zou, "EDA: Easy data augmentation techniques for boosting performance on text classification tasks," in *Proceedings of the 2019 Conference on Empirical Methods in Natural Language Processing and the 9th International Joint Conference on Natural Language Processing*, pp. 6382–6388. Nov. 2019.

[9] H. Zhao, H. Chen, T. A. Ruggles, Y. Feng, D. Singh, and H.-J. Yoon, "Improving text classification with large language model-based data augmentation," *Electronics (Basel)*, vol. 13, no. 13, p. 2535, June. 2024.

[10] L. Becker, P. Pracht, P. Sertdal, J. Uboreck, A. Bendel, and R. Martin, "Conditional label smoothing for LLM-based data augmentation in medical text classification," in *Proceedings of the 2024 IEEE Spoken Language Technology Workshop*, pp. 833–840, Jan. 2025.

[11] S. Yang, X. Liu, X. Dong, and B. Fu, "Mini-DA: Improving your model performance through minimal data augmentation using LLM," in *Proceedings of the Fifth Workshop on Data Science with Human-in-the-Loop*, pp. 25–30, Jun. 2024.

[12] S. V. Balkus and D. Yan, "Improving short text classification with augmented data using GPT-3," *Natural Language Engineering*, vol. 30, no. 5, pp. 943–972, Sep. 2024.

[13] H. Dai et al., "AugGPT: Leveraging ChatGPT for Text Data Augmentation," *IEEE Transactions on Big Data*, vol. 11, no. 3, pp. 907–918, June. 2025.

[14] Y. Chai, H. Xie, and J. S. Qin, "Text data augmentation for large language models: A comprehensive survey of methods, challenges, and opportunities," *arXiv [cs.CL]*, Jan. 2025.

[15] D. Zhang, R. Mi, P. Zhou, D. Jin, M. Zhang, and T. Song, "Large model-based data augmentation for imbalanced text classification," in *Proceedings of the 2024 5th International Seminar on Artificial Intelligence, Networking and Information Technology*, pp. 1006–1010, Jul. 2024.

[16] N. V. Chawla, K. W. Bowyer, L. O. Hall, and W. P. Kegelmeyer, "SMOTE: Synthetic minority over-sampling technique," *Journal of Artificial Intelligence Research*, vol. 16, pp. 321–357, June. 2002.

[17] L. Wang et al., "Prompt engineering in consistency and reliability with the evidence-based guideline for LLMs," *NPJ Digital Medicine*, vol. 7, no. 1, p. 41, Feb. 2024.

[18] H. Liu, H. Yin, Z. Luo, and X. Wang, "Integrating chemistry knowledge in large language models via prompt engineering," *Synthetic and Systems Biotechnology*, vol. 10, no. 1, pp. 23–38, Mar. 2025.

[19] R. Feng et al., "Engineering of generative artificial intelligence and natural language processing models to accurately identify arrhythmia recurrence," *Circulation: Arrhythmia and Electrophysiology*, vol. 18, no. 1, p. e013023, Jan. 2025.

[20] S. Morales, R. Clarisó, and J. Cabot, "Impromptu: a framework for model-driven prompt engineering," *Software and Systems Modeling*, pp. 1–19, Jan. 2025.

[21] K. Zhou, J. Yang, C. C. Loy, and Z. Liu, "Learning to prompt for vision-language models," *International Journal of Computer Vision*, vol. 130, no. 9, pp. 2337–2348, Sept. 2022.

[22] H. Wei and Z. Chen, "Improving domain generalization for Image Captioning with unsupervised prompt learning," *ACM Transactions on Multimedia Computing, Communications, and Applications*, no. 3715136, Feb. 2025.

[23] J. Ye, Z. Wu, J. Feng, T. Yu, and L. Kong, "Compositional Exemplars for in-context Learning," in *Proceedings of the 40th International Conference on Machine Learning*, no. 1662, pp.39818–39833, Jul. 2023.

[24] Z. Yang, Y. Zhang, D. Sui, C. Liu, J. Zhao, and K. Liu, "Representative demonstration selection for in-context learning with two-stage determinantal point process," in *Proceedings of the 2023 Conference on Empirical Methods in Natural Language Processing*, pp. 5443–5456, Dec. 2023.

[25] J. Kapuriya, M. Kaushik, D. Ganguly, and S. Bhatia, "Exploring the role of diversity in example selection for in-context learning," in *Proceedings of the 48th International ACM SIGIR Conference on Research and Development in Information Retrieval*, pp. 2962–2966, Jul. 2025.

[26] S. A. Hayati, M. Lee, D. Rajagopal, and D. Kang, "How far can we extract diverse perspectives from large language models?," in *Proceedings of the 2024 Conference on Empirical Methods in Natural Language Processing*, pp. 5336–5366, Nov. 2024.

[27] Q. Wang, S. Pan, T. Linzen, and E. Black, "Multilingual prompting for improving LLM generation diversity," in *Proceedings of the 2025 Conference on Empirical Methods in Natural Language Processing*, pp. 6378–6400, Dec. 2025.

[28] W. Hu, G. K. R. Lau, L. Diwen, C. Jizhuo, S.-K. Ng, and B. K. H. Low, "Dipper: Diversity in prompts for producing large language model ensembles in reasoning tasks," in *Proceedings of the 2025 Conference on Empirical Methods in Natural Language Processing*, pp. 35546–35560, Dec. 2025.